\documentclass[letterpaper, 10 pt, conference]{ieeeconf}  

\IEEEoverridecommandlockouts                              
\usepackage{graphicx}
\usepackage{graphics} 
\usepackage[normalem]{ulem}
\usepackage{array}
\usepackage{hyperref}
\usepackage{subcaption}
\useunder{\uline}{\ul}{}
\usepackage{multirow}

\title{\LARGE \bf
	Anytime Stereo Image Depth Estimation on Mobile Devices
}
\newcommand{\namelong}[1]{Anytime Stereo Network}
\newcommand{\nameshort}[1]{AnyNet}
\newcommand{\spnlong}[1]{Spatial Propagation Network}
\newcommand{\spnshort}[1]{SPNet}
\newcolumntype{P}[1]{>{\centering\arraybackslash}p{#1}}
\newcolumntype{M}[1]{>{\centering\arraybackslash}m{#1}}

\author{Yan Wang$^{*1}$, Zihang Lai$^{*2}$, Gao Huang$^{1}$, Brian H. Wang$^{1}$, Laurens van der Maaten$^{3}$, \\ Mark Campbell$^{1}$, and Kilian Q. Weinberger$^{1}$
\thanks{* Authors contributed equally}
\thanks{1 Cornell University. \{yw763, gh349, bhw45, mc288, kqw4\}@cornell.\newline edu}
\thanks{2 University of Oxford. This work performed during Zihang Lai's internship at Cornell University.
	{\small zihang.lai@cs.ox.ac.uk}}
\thanks{3 Facebook AI Research.
	{\small lvdmaaten@gmail.com}}
 }

\begin{document}

	\maketitle
	\thispagestyle{empty}
	\pagestyle{empty}
	
	\begin{abstract}		
  %
  Many applications of stereo depth estimation in robotics require the generation of accurate disparity maps in real time under significant computational constraints. Current state-of-the-art algorithms force a choice between either generating accurate mappings at a slow pace, or quickly generating inaccurate ones, and additionally these methods typically require far too many parameters to be usable on power- or memory-constrained devices. 
  Motivated by these shortcomings, we propose a novel approach for \emph{disparity prediction} in the \emph{anytime} setting. In contrast to prior work, our end-to-end learned approach can trade off computation and accuracy at inference time. Depth estimation is performed in stages, during which the model can be queried at any time to output its current best estimate.  
  Our final model can process 1242$ \times $375 resolution images within a range of 10-35 FPS on an NVIDIA Jetson TX2 module with only marginal increases in error -- using two orders of magnitude fewer parameters than the most competitive baseline. 
 The source code is available at \url{https://github.com/mileyan/AnyNet}.
	\end{abstract}


\section{INTRODUCTION}

Depth estimation from stereo camera images is an important task for 3D scene reconstruction and understanding, with numerous applications ranging from robotics~\cite{mancini2016fast,ye2017self,saxena2007depth,schmid2013stereo} to augmented reality \cite{zenati2007dense,alhaija2018augmented,nguyen2018depth}. High-resolution stereo cameras provide a reliable solution for 3D perception - unlike time-of-flight cameras, they work well both indoors and outdoors, and compared to LiDAR they are substantially more affordable and energy-efficient \cite{luo2016efficient}. Given a rectified stereo image pair, the focal length, and the stereo baseline distance between the two cameras, depth estimation can be cast into a stereo matching problem, the goal of which is to find the disparity between corresponding pixels in the two images. Although disparity estimation from stereo images is a long-standing problem in computer vision~\cite{lucas1981iterative}, in recent years the adoption of deep convolutional neural networks (CNN)~\cite{zbontar2016stereo,mayer2016large,kendall2017end,liang2017learning,pang2017cascade} has led to significant progress in the field. Deep networks can solve the matching problem via supervised learning in an end-to-end fashion, and they have the ability to incorporate local context as well as prior knowledge into the estimation process. 

\begin{figure}[t]
  \centering
  \includegraphics[width=\columnwidth]{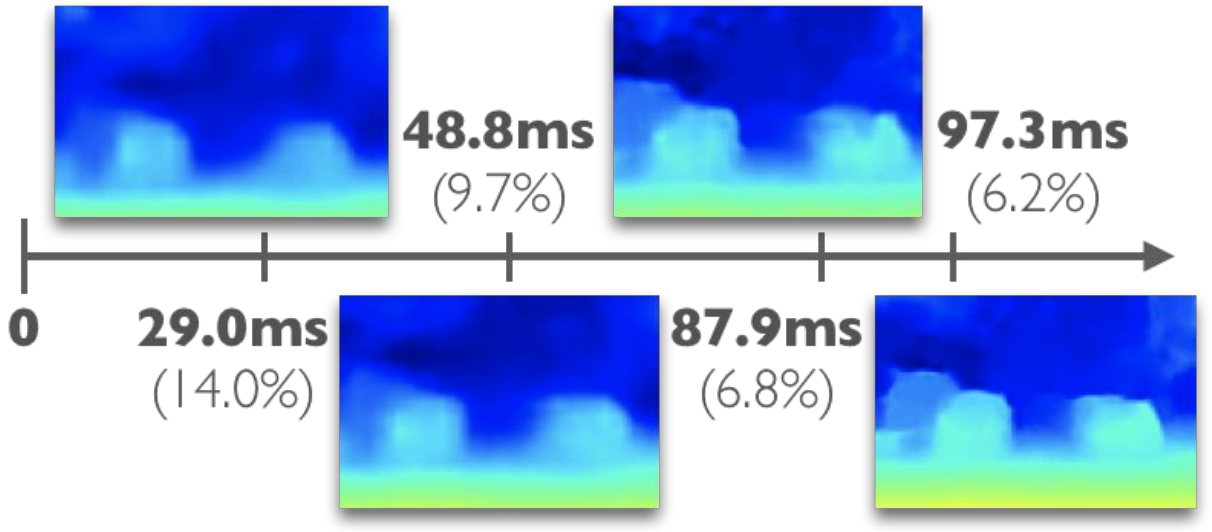}
\caption{Example timeline of \nameshort{} predictions. As time progresses the depth estimation becomes increasingly accurate. The algorithm can be polled at any time to return the current best estimate of the depth map. The initial estimates may be sufficient to trigger an obstacle avoidance maneuver, whereas the later images contain enough detail for more advanced path planning procedures. (3-pixel error rate below time.)  }
  \label{fig:timeline}
 \vspace{-0.4cm}
\end{figure}

On the other hand, deep neural networks tend to be computationally intensive and suffer from significant latency when processing high-resolution stereo images. For example, PSMNet~\cite{zhao2017pyramid}, arguably the current state-of-the-art algorithm for depth estimation, obtains a frame rate below 0.3FPS on the Nvidia Jetson TX2 GPU computing module --- far too slow for timely obstacle avoidance by drones or other autonomous robots. 

In this paper, we argue for an \emph{anytime} computational approach to disparity estimation, and present a model that trades off between speed and accuracy dynamically (see Figure~\ref{fig:timeline}). For example, an autonomous drone flying at high speed can poll our 3D depth estimation method at a high frequency. If an object appears in its flight path, it will be able to perceive it rapidly and react accordingly by lowering its speed or performing an evasive maneuver. When flying at low speed, latency is not as detrimental, and the same drone could compute a higher resolution and more accurate 3D depth map, enabling tasks such as high precision navigation in crowded scenes or detailed mapping of an environment. 

The computational complexity of depth estimation with convolutional networks typically scales cubically with the image resolution, and linearly with the maximum disparity that is considered~\cite{kendall2017end}. Keeping these characteristics in mind, we refine the depth map successively, while always ensuring that either the resolution or the maximum disparity range is sufficiently low to ensure minimal computation time. 
We start with low resolution ($1/16$) estimates of the depth map at the full disparity range. The cubic complexity allows us to compute this initial depth map in a few milliseconds (where the bulk of the time is spent on the initial feature extraction and down-sampling). Starting with this low resolution estimate, we successively increase the resolution of the disparity map by up-sampling and subsequently correcting the errors that are now apparent at the higher resolution. Correction is performed by predicting the residual error of the up-sampled disparity map from the input images with a CNN. 
Despite the higher resolution used, these updates are still fast because the residual disparity can be assumed to be bounded within a few pixels, allowing us to restrict the maximum disparity, and associated computation, to a mere $10-20\%$ of the full range.

These successive updates avoid full-range disparity computation at all but the initial low resolution setting, and ensure that all computation is re-used, setting our method apart from most existing multi-scale network structures~\cite{saxena2016convolutional,huang2017multi,ke2016neural}. Furthermore, our algorithm can be polled at any time in order to retrieve the current best estimated depth map. A wide range of possible frame rates are attainable (10-35FPS on a TX2 module), while still preserving accurate disparity estimation in the high-latency setting. Our entire network can be trained end-to-end using a joint loss over all scales, and we  refer to it as \namelong{} (\nameshort{}).

We evaluate \nameshort{} on multiple benchmark data sets for depth estimation, with various encouraging findings: Firstly, \nameshort{} obtains competitive accuracy with state of the art approaches, while having orders of magnitude fewer parameters than the baselines. This is especially impactful for resource-constrained embedded devices. Secondly, we find that deep convolutional networks are highly capable at predicting residuals from coarse disparity maps. Finally, including a final spatial propagation model (SPNet)~\cite{liu2017learning} significantly improves the disparity map quality, yielding state-of-the-art results at a fraction of the computational cost (and parameter storage requirements) of existing methods.




\section{RELATED WORK}
\paragraph{\textbf{Disparity estimation}} Traditional approaches to disparity estimation are based on matching features between the left and right images \cite{barnard1982computational,scharstein2002taxonomy}. These approaches are typically comprised of the following four steps: (1) computing the costs of matching image patches over a range of disparities, (2) smoothing the resulting cost tensor via aggregation methods, (3) estimating the disparity by finding a low-cost match between the patches in the left image and those in the right image, and (4) refining these disparity estimates by introducing global smoothness priors on the disparity map \cite{scharstein2002taxonomy,hosni2013,hirschmuller2009,hu2012stereo}. Several recent papers have studied the use of convolutional networks in step (1). In particular, Zbontar \& LeCun \cite{zbontar2016stereo} use a Siamese convolutional network to predict patch similarities for matching left and right patches. Their method was further improved via the use of more efficient matching networks \cite{luo2016efficient} and deeper highway networks trained to minimize a multilevel loss \cite{shaked2017improved}.

\paragraph{\textbf{End-to-end disparity prediction}} Inspired by these initial successes of convolutional networks in disparity estimation, as well as by their successes in semantic segmentation \cite{long2015fully}, optical flow computation \cite{fischer2015flownet,ilg2017flownet}, and depth estimation from a single frame \cite{eigen2015single}, several recent studies have explored end-to-end disparity estimation models \cite{mayer2016large,kendall2017end,liang2017learning,pang2017cascade}. For example, in \cite{mayer2016large}, the disparity prediction problem is formulated as a supervised learning problem, and a convolutional network called DispNet is proposed that directly predicts disparities for an image pair. Improvements made by DispNet include a cascaded refinement procedure \cite{pang2017cascade}. Other studies adopt the \emph{correlation layer} introduced in \cite{fischer2015flownet} to obtain the initial matching costs; a set of two convolutional networks are trained to predict and further refine the disparity map for the image pair \cite{liang2017learning}. Several prior studies have also explored moving away from the supervised learning paradigm by performing depth estimation in an unsupervised fashion using stereo images \cite{godard2017unsupervised} or video streams \cite{zhou2017unsupervised}.

Our work is partially inspired by the Geometry and Context Network (GCNet) proposed in \cite{kendall2017end}. In order to predict the disparity map between two images, GCNet combines a 2D Siamese convolutional network operating on the image pair with a 3D convolutional network that operates on the matching cost tensor. GCNet is trained in an end-to-end fashion, and is presently one of the state-of-the-art methods in terms of accuracy and computational efficiency. Our model is related to GCNet in that it has the same two stages (a 2D Siamese image convolutional network and a 3D cost tensor convolutional network), but it differs in that it can perform \emph{anytime prediction} by rapidly producing an initial disparity map prediction and then progressively predicting the residual to correct this prediction. LiteFlowNet~\cite{hui18liteflownet} also tries to predict the residual for optical flow computation. However, LiteFlowNet uses the residual to facilitate large-displacement flow inference rather than computational speedup.

\paragraph{\textbf{Anytime prediction}} There exists a substantial body of work on machine learned models with computational budget constraints at inference time \cite{viola2001robust,grubb2012speedboost,karayev2014anytime,greedymiser,xu2013cost,wang2015acyclic,huang2017multi}. Most of these approaches are based on ensembles of decision-tree classifiers~\cite{viola2001robust,greedymiser,grubb2012speedboost,xu2013cost} which allow for tree-by-tree evaluation, facilitating the progressive prediction updates that are the hallmark of anytime prediction. Several recent studies have explored anytime prediction with image classification CNNs that dynamically evaluate parts of the network to progressively refine their predictions \cite{fractalnet,huang2017multi,veit2017,wu2017blockdrop,mcgill2017,figurnov2017}. Our work differs from these earlier anytime CNN models in that we focus on the structured prediction problem of disparity-map estimation, rather than on image classification tasks. Our models exploit particular properties of the disparity-prediction problem: namely, that progressive estimation of disparities can be achieved by progressively increasing the resolution of the image data within the internal representation of the CNN.


\section{\nameshort{}}

\begin{figure*}[ht]
  \centering
  \includegraphics[width=0.7\paperwidth]{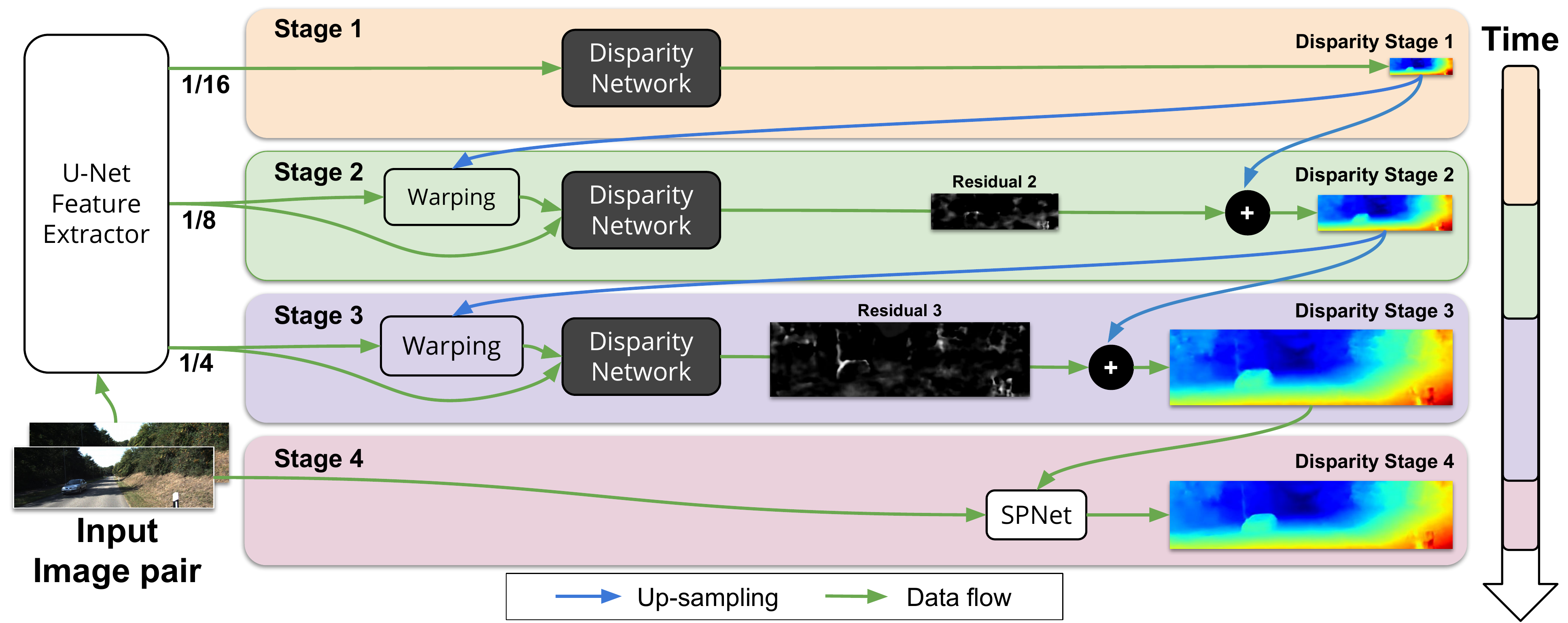}
  \caption{Network structure of \nameshort{}.}
  \label{fig:warp}
\end{figure*}

\begin{figure}[t]
  \centering
  \includegraphics[width=\columnwidth]{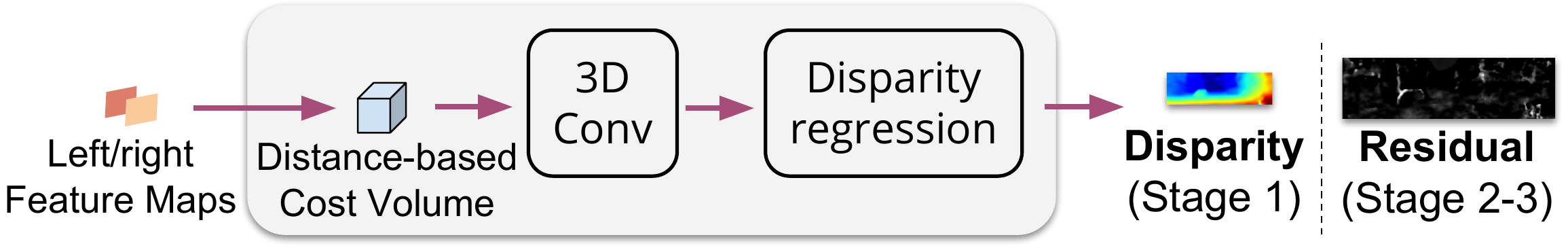}
\caption{Disparity network. See text for details.}
  \label{fig:disp}
  \vspace{-0.5cm}
\end{figure}

Fig.~\ref{fig:warp} shows a schematic layout of the \nameshort{} architecture. An input image pair first passes through the U-Net feature extractor, which computes feature maps at several output resolutions (of scale 1/16, 1/8, 1/4). In the first stage, only the lowest-scale features (1/16) are computed and passed through a disparity network (Fig.~\ref{fig:disp}) to produce a low-resolution disparity map (\emph{Disparity Stage 1}). A disparity map estimates the horizontal offset of each pixel in the right input image w.r.t. the left input image, and can be used to compute a depth map. 
Because of the low input resolution, the entire Stage 1 computation requires only a few milliseconds. If more computation time is permitted, we enter Stage 2 by continuing the computation in the U-Net to obtain larger-scale (1/8) features. 
Instead of computing a full disparity map at this higher resolution, in Stage 2 we simply correct the already-computed disparity map from Stage 1.  
First, we up-scale the disparity map to match the resolution of Stage 2. We then compute a residual map, which contains small corrections that specify how much the disparity map should be increased or decreased for each pixel.  
If time permits, Stage 3 follows a similar process as Stage 2, and doubles the resolution again from a scale of 1/8 to 1/4. Stage 4 refines the disparity map from Stage 3 with an SPNet~\cite{liu2017learning}.

In the remainder of this section we describe the individual components of our model in greater detail.

\begin{figure}[t]
  \centering
  \includegraphics[width=\columnwidth]{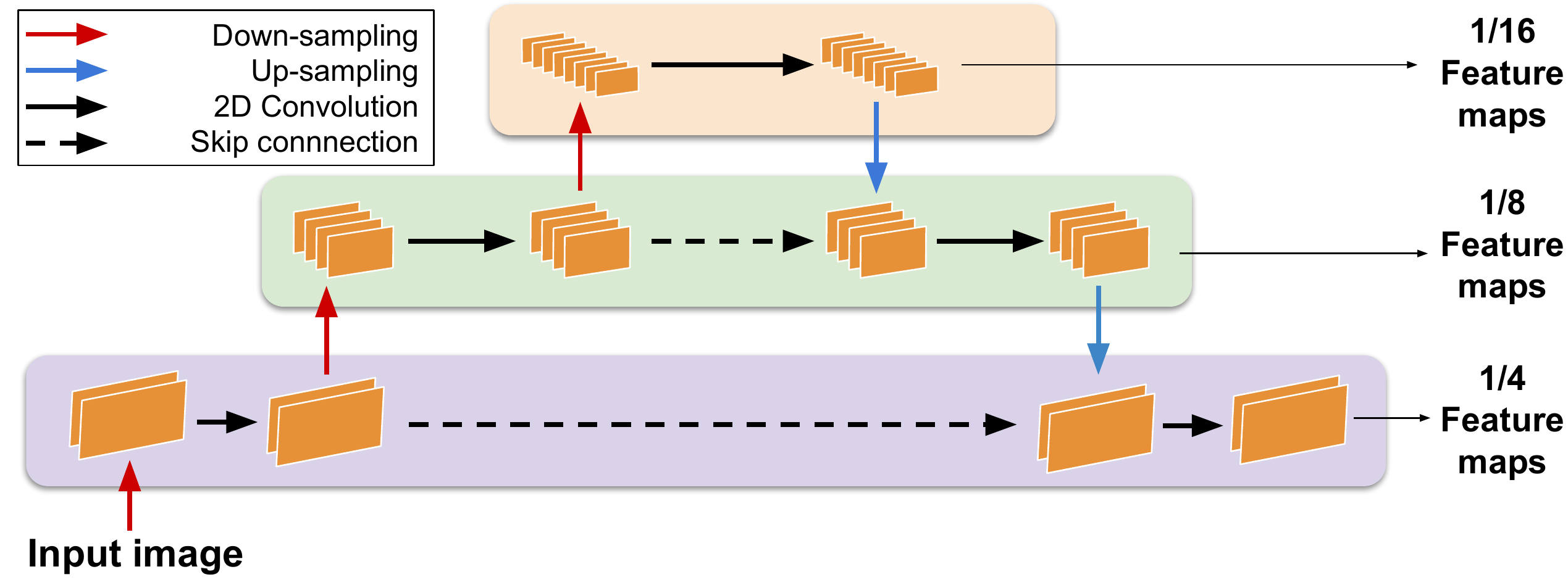}

\caption{U-Net Feature Extractor. See text for details.}
  \label{fig:fe}
\vspace{-0.5cm}
\end{figure}
\paragraph{U-Net Feature Extractor}
 Fig~\ref{fig:fe} illustrates the U-Net~\cite{ronneberger2015u} Feature Extractor in detail, which is applied to  both the left and right image. The U-Net architecture computes feature maps at various resolutions (1/16, 1/8, 1/4), which are used as input at stages 1-3 and only computed when needed. The original input images are down-sampled through max-pooling or strided convolution and then processed with convolutional filters. Lower resolution feature maps capture the global context, whereas higher resolution feature maps capture local details. At scale 1/8 and 1/4, the final convolutional layer incorporates the previously computed lower-scale features. 


\paragraph{Disparity Network}
The disparity network (Fig.~\ref{fig:disp}) takes as input the feature maps from the left and right stereo images in order to compute a disparity map. We use this component to compute the initial disparity map (stage 1) as well as to compute the residual maps for subsequent corrections (stages 2 \& 3). The disparity network first computes a disparity cost volume. Here, the cost refers to the similarity between a pixel in the left image and a corresponding pixel in the right image.  If the input feature maps are of dimensions $H\times W$, the cost volume has dimensions $H\times W \times M$, where the $(i,j,k)$ entry describes the degree to which pixel $(i,j)$ of the left image matches pixel $(i,j\!-\!k)$ in the right image. $M$ denotes the maximum disparity under consideration. We can represent each pixel $(i,j)$ in the left image as a vector $\mathbf{p}_{ij}^L$, where dimension $\alpha$ corresponds to the $(i,j)$ entry in the $\alpha^{th}$ input feature map  associated with the left image. Similarly we can define $\mathbf{p}_{ij}^R$. The entry $(i,j,k)$ in the cost volume is then defined as the $L_1$ distance between the two vectors $\mathbf{p}_{ij}^L$ and $\mathbf{p}_{i(j+k)}^R$, i.e. $C_{ijk}=\|\mathbf{p}_{ij}^L-\mathbf{p}_{i(j\!-\!k)}^R\|_1$.

This cost volume may still contain errors due to blurry objects, occlusions, or ambiguous matchings in the input images. As a second step in the disparity network (\emph{3D Conv} in Fig~\ref{fig:disp}), we refine the cost volume with several 3D convolution layers~\cite{kendall2017end} to further improve the obtained cost volume.


The disparity for pixel $(i,j)$ in the left image is $k$ if the pixel $(i,j\!-k\!)$ in the right image is most similar. If the cost volume is exact, we could therefore compute the disparity of pixel $(i,j)$ as $\hat{D}_{ij}=\arg\!\min_{k} C_{i,j,k}$.  However, the cost estimates may be too noisy to search for the hard minimum. Instead, we follow the suggestion by Kendall et al.~\cite{kendall2017end} and compute a weighted average  (\emph{Disparity regression} in Fig~\ref{fig:disp})
\begin{equation}
\hat{D}_{ij} = \sum_{k=0}^{M}k\times \frac{\exp{(-C_{ijk})} } {\sum_{k'=0}^{M}\exp{(-C_{ijk'})}}.
\end{equation}
If one disparity $k$ clearly has the lowest cost (i.e. it is the only good match), it will be recovered by the weighted average. If there is ambiguity, the output will be an average of the viable candidates. 



\paragraph{Residual Prediction}
A crucial aspect of our \nameshort{} architecture is that we only compute the full disparity map at a very low resolution in Stage 1. In Stages 2 \& 3 we predict residuals~\cite{hui18liteflownet}. The most expensive part of the disparity prediction is the construction and refinement of the cost volume. 
The cost volume scales $H\times W\times M$, where $M$ is the maximum disparity. In high resolutions, the maximum disparity between two pixels can be very large (typically $M=192$ pixels in the KITTI dataset~\cite{Geiger2012CVPR}). 
By restricting ourselves to residuals, i.e. corrections of existing disparities, we can limit ourselves to $M=5$ (corresponding to offsets $-2,-1,0,1,2$) and obtain sizable speedups. 

In order to compute residuals in stages 2 \& 3, we first up-scale the coarse disparity map and use it to warp the input features at the higher scale (Fig.~\ref{fig:warp}) by applying the disparity estimations pixel-wise. In particular, if the left disparity of pixel $(i,j)$ is estimated to be $k$, we overwrite the value of pixel $(i,j)$  in each right feature map to the corresponding value of pixel $(i,j+k)$ (using zero if out of bounds). If the current disparity estimate is correct, the updated right feature maps should match the left feature maps. Due to the coarseness of the low resolution inputs, there is typically still a mismatch of several pixels, which we correct by computing residual disparity maps. 
Prediction of the residual disparity is accomplished similarly to the full disparity map computation. The only difference is that the cost volume is computed as $C_{ijk}=\|\mathbf{p}_{ij}-\mathbf{p}_{i(j-k+2)}\|_1$, and the resulting residual disparity map is added to the up-scaled disparity map from the previous stage.

\paragraph{\spnlong{}}
To further improve our results, we add a final fourth stage in which we use a \spnlong{} (\spnshort{})~\cite{liu2017learning} to refine our disparity predictions. The \spnshort{} sharpens the disparity map by applying a local filter whose weights are predicted by applying a small CNN to the left input image. We show that this refinement improves our results significantly at relatively little extra cost.

\begin{table}[h!]
\centering
\scriptsize
\begin{tabular}{l|l}
 \hline
 0 & Input image  \\
 \hline
 \multicolumn{2}{c} { \textbf{2-D Unet features} } \\
 \hline
  1 & $3 \!\times\! 3$ conv with 1 filter \\
  2 & $3 \!\times\! 3$ conv with stride 2 and 1 filter \\
  3 & $2 \!\times\! 2$ maxpooling with stride 2  \\
4-5 & $3 \!\times\! 3$ conv with 2 filters \\
  6 & $2 \!\times\! 2$ maxpooling with stride 2  \\
7-8 & $3 \!\times\! 3$ conv with 4 filters \\
  9 & $2 \!\times\! 2$ maxpooling with stride 2  \\
10-11 & $3 \!\times\! 3$ conv with 8 filters  \\
 12 & Bilinear upsample layer 11 (features) into 2x size  \\
 13 & Concatenate layer 8 and 12  \\
14-15 & $3 \!\times\! 3$ conv with 4 filters \\
 16 & Bilinear upsample layer 15 (features) into 2x size  \\
 17 & Concatenate layer 5 and 16  \\
18-19 & $3 \!\times\! 3$ conv with 2 filters \\ \hline
 \multicolumn{2}{c} { \textbf{Cost volume} } \\
 \hline
 20 & Warp and build cost volume from layer 11 \\
 21 & Warp and build cost volume from layer 15 and layer 29  \\
 22 & Warp and build cost volume from layer 19 and layer 36  \\
 \hline
 \multicolumn{2}{c} { \textbf{Regularization} } \\
 \hline
 23-27 & $3 \!\times\! 3 \!\times\! 3$ 3-D conv with 16 filters \\
 28 & $3 \!\times\! 3 \!\times\! 3$ 3-D conv with 1 filter \\
 29 & Disparity regression \\
 30 & Upsample layer 29 to image size: \textbf{stage 1 disparity output}   \\
 31-35 & $3 \!\times\! 3 \!\times\! 3$ 3-D conv with 4 filters \\
 36 & Disparity regression: residual of stage 2 \\
 37 & Upsample 36 it into image size   \\
 38 & Add layer 37 and layer 30 \\
 39 & Upsample layer 38 to image size: \textbf{stage 2 disparity output} \\
 40-44 & $3 \!\times\! 3 \!\times\! 3$ 3-D conv with 4 filters \\
 45 & Disparity regression: residual of stage 3 \\
 46 & Add layer 44 and layer 38   \\ 
 47 & Upsample layer 46 to image size: \textbf{stage 3 disparity output}  \\\hline
 \multicolumn{2}{c} { \textbf{Spatial propagation network} } \\
 \hline
 48-51 & $3 \!\times\! 3$ conv with 16 filters (on input image)\\ 
 52 & $3 \!\times\! 3$ conv with 24 filters: affinity matrix\\ 
 53 & $3 \!\times\! 3$ conv with 8 filters (on layer 47)\\ 
 54 & Spatial propagate layer 53 with layer 52 (affinity matrix) \\
 55 & $3 \!\times\! 3$ conv with 1 filters: \textbf{stage 4 disparity output}  \\ 
\hline

\end{tabular}
\caption{Network configurations. Note that a \emph{conv} stands for a sequence of operations: batch normalization, rectified linear units (ReLU) and convolution. The default stride is 1.\label{table:network-detail}}

\end{table}
	

\section{EXPERIMENTAL RESULTS}
In this section, we empirically evaluate our method and compare it with existing stereo algorithms. In addition, we benchmark the efficiency of our approach on an Nvidia Jetson TX2 computing module. 

\begin{figure}[t]
  \centering
 \includegraphics[width=0.8\linewidth ]{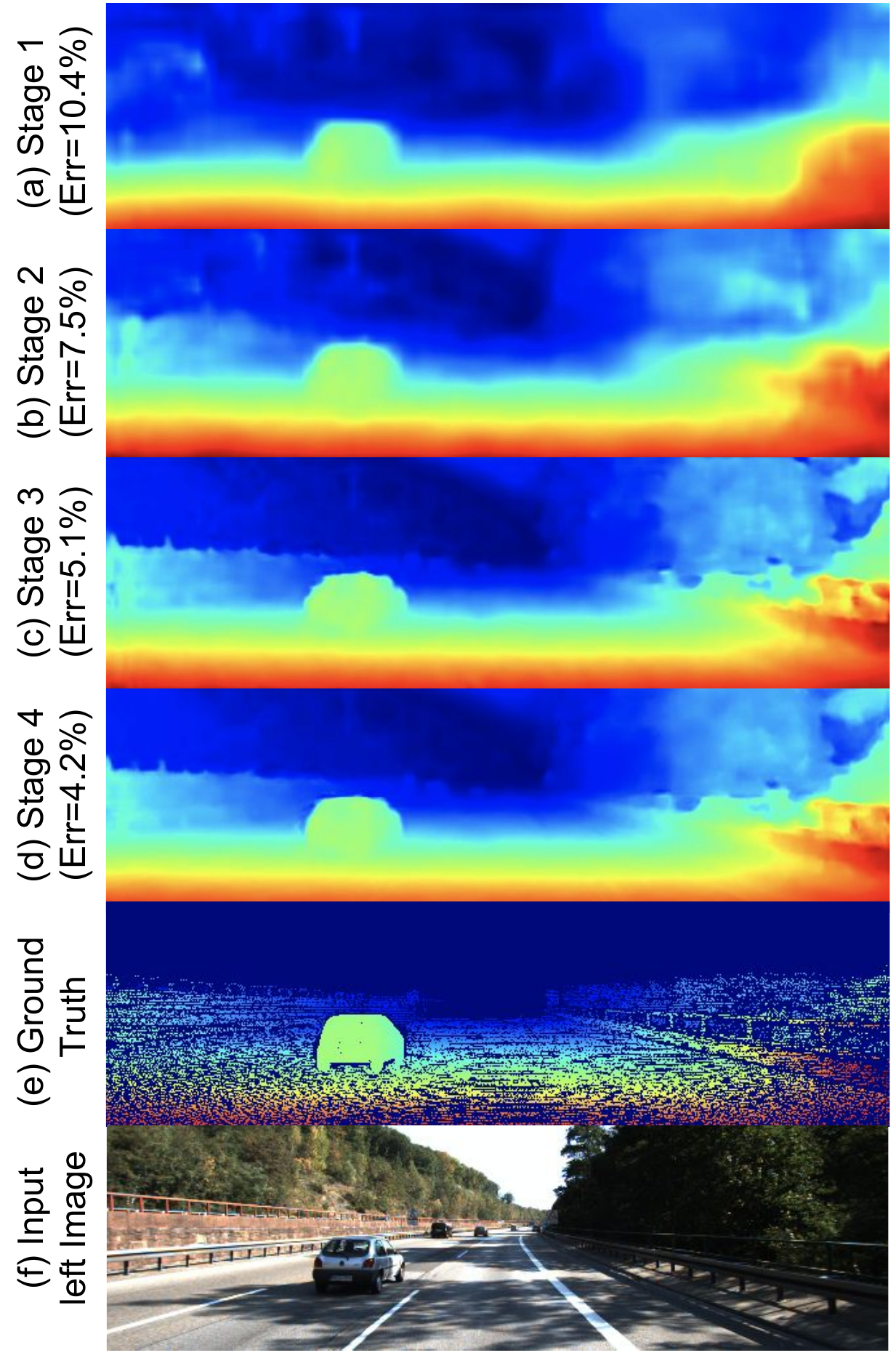}
\caption{(a)-(d) Disparity prediction from 4 stages of \nameshort{} on KITTI-2015. As a larger computational budget is made available, the prediction is refined and becomes more accurate. (e) shows the ground truth LiDAR image, and (f) shows the left input image.}
  \label{fig:vis}
  \vspace{-0.5cm}
\end{figure}

 \paragraph{\textbf{Implementation Details}}
We implement \nameshort{} in PyTorch~\cite{paszke2017automatic}. See Table~\ref{table:network-detail} for a detailed network description.
Our experiments use an \nameshort{} implementation with four stages, as shown in Figure~\ref{fig:warp} and described in the previous section. The maximum disparity is set to $192$ pixels in the original image, which corresponds to a Stage 1 cost volume depth of $M=192/16=12$. In Stages 2 \& 3 the residual range is $\pm 2$, corresponding to $\pm 16$ pixels in Stage 2 and $\pm 8$ pixels in Stage 3.
All four stages, including the \spnshort{} in Stage 4, are trained jointly, but the losses are weighted differently, with weights $\lambda_{1}=1/4$, $\lambda_{2}=1/2$, $\lambda_{3}=1$ and $\lambda_{4}=1$, respectively.  In total, our model contains 40,000 parameters - this is an order of magnitude fewer parameters than StereoNet~\cite{khamis2018stereonet}, and two orders of magnitude fewer than PSMNet~\cite{zhao2017pyramid}. Our model is trained end-to-end using Adam~\cite{kingma2014adam} with initial learning rate $5e^{-4}$ and batch size 6. On the Scene Flow dataset~\cite{mayer2016large}, the learning rate is kept constant, and the training lasts for 10 epochs in total. For the KITTI dataset we first pre-train the model on Scene Flow, before fine-tuning it for 300 epochs.
The learning rate is divided by 10 after epoch 200. All input images are normalized to be zero-mean with unit variance. All experiments were conducted using original image resolutions. Using one GTX 1080Ti GPU, training on the Scene Flow dataset took 3.5 hours, and training on KITTI took 30 minutes. All results are averaged over five randomized 80/20 train/validation splits. 

 Figure~\ref{fig:vis} visualizes the disparity maps predicted at the four stages of our model. As more computation time is made available, \nameshort{} produces increasingly refined disparity maps. The final output from stage 4 is even sharper and more accurate, due to the SPNet post-processing.

\paragraph{\textbf{Datasets}} Our model is trained on the synthetic Scene Flow \cite{mayer2016large} dataset and evaluated on two real-world datasets, KITTI-2012~\cite{Geiger2012CVPR} and KITTI-2015~\cite{Menze2015CVPR}. The Scene Flow dataset contains $22,000$ stereo image pairs for training, and $4,370$ image pairs for testing. Each image has a resolution of $960\times 540$ pixels. As in \cite{kendall2017end}, we train our model on $512\times 256$ patches randomly cropped from the original images. The KITTI-2012 dataset contains 194 pairs of images for training and 195 for testing, while KITTI-2015 contains 200 image pairs for each. All of the KITTI images are of size $1242 \times 375$.

\paragraph{\textbf{Baselines}} 
Although state-of-the-art CNN based stereo estimation methods have been reported to reach 60FPS on a TITAN X GPU~\cite{khamis2018stereonet}, they are far from achieving real-time performance on more resource-constrained computing devices such as the Nvidia Jetson TX2.
Here, we present a controlled comparison on a TX2 between our method and four competitive baseline algorithms: PSMNet~\cite{zhao2017pyramid}, StereoNet~\cite{khamis2018stereonet}, DispNet~\cite{mayer2016large}, and StereoDNN~\cite{smolyanskiy2018importance}. The PSMNet model has two different versions: PSMNet-classic and PSMNet-hourglass. We use the former, as it is much more efficient than PSMNet-hourglass while having comparable accuracy. For StereoNet, we report running times using a Tensorflow implementation, which we found to be twice as fast as a PyTorch implementation.

Finally, we also compare \nameshort{} to two classical stereo matching approaches: Block Matching~\cite{konolige1998small} and Semi-Global Block Matching~\cite{hirschmuller2008stereo}, supported by OpenCV~\cite{bradski2000opencv}.

In order to collect meaningful results for these baseline methods on the TX2, we use down-sampled input images for faster inference times. The baseline methods are re-implemented, and trained on down-sampled stereo images - this allows a fair comparison, since a model trained on full-sized images would be expected to suffer a significant performance decrease when given lower-resolution inputs. After obtaining a low-resolution prediction, we up-sample it to the original size using bilinear interpolation.

\subsection{Evaluation Results}
\begin{figure}
  \centering
  \vspace{-0.5cm}
  \begin{subfigure}[t]{0.8\linewidth}
   \includegraphics[width=\linewidth ]{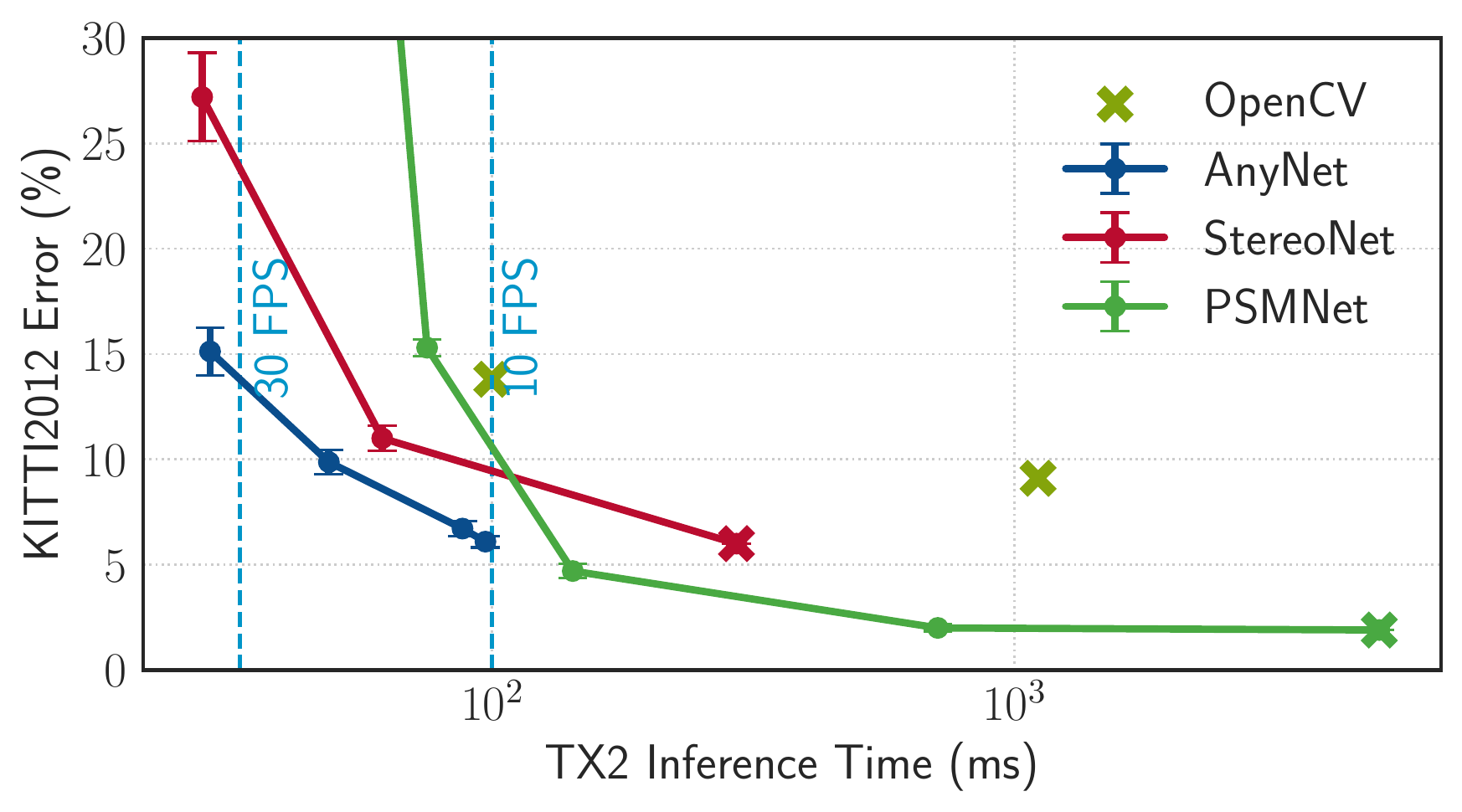}
  \caption{KITTI-2012 results across different down-sampling resolutions}
  \label{fig:2012_main_result_log_scale}
  \end{subfigure}\hfill
  \begin{subfigure}[t]{0.8\linewidth}
     \includegraphics[width=\linewidth ]{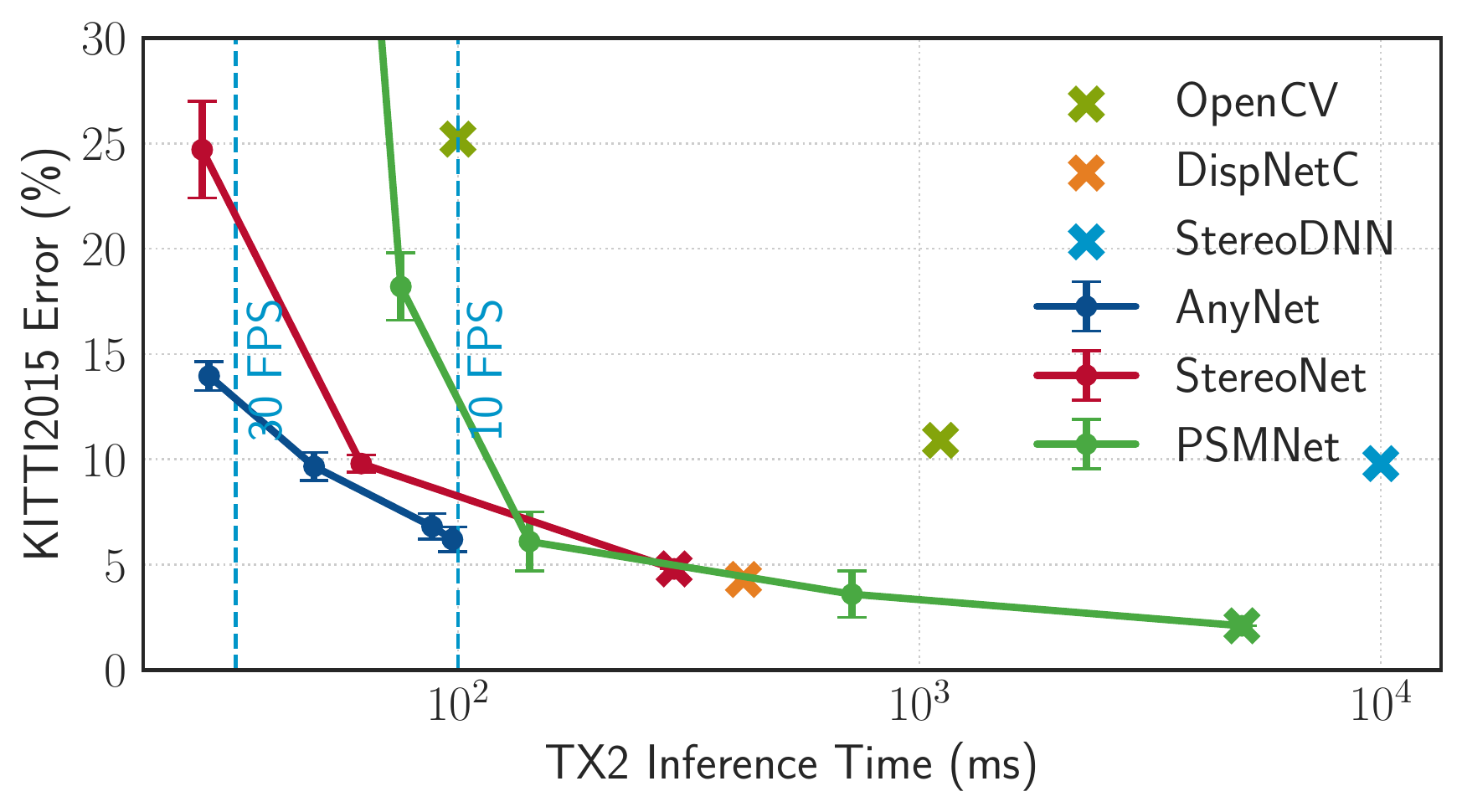}
    \caption{KITTI-2015 results across different down-sampling resolutions}
      \label{fig:2015_main_result_log_scale}
  \end{subfigure}
  \begin{subfigure}[t]{0.8\linewidth}

   \includegraphics[width=\linewidth ]{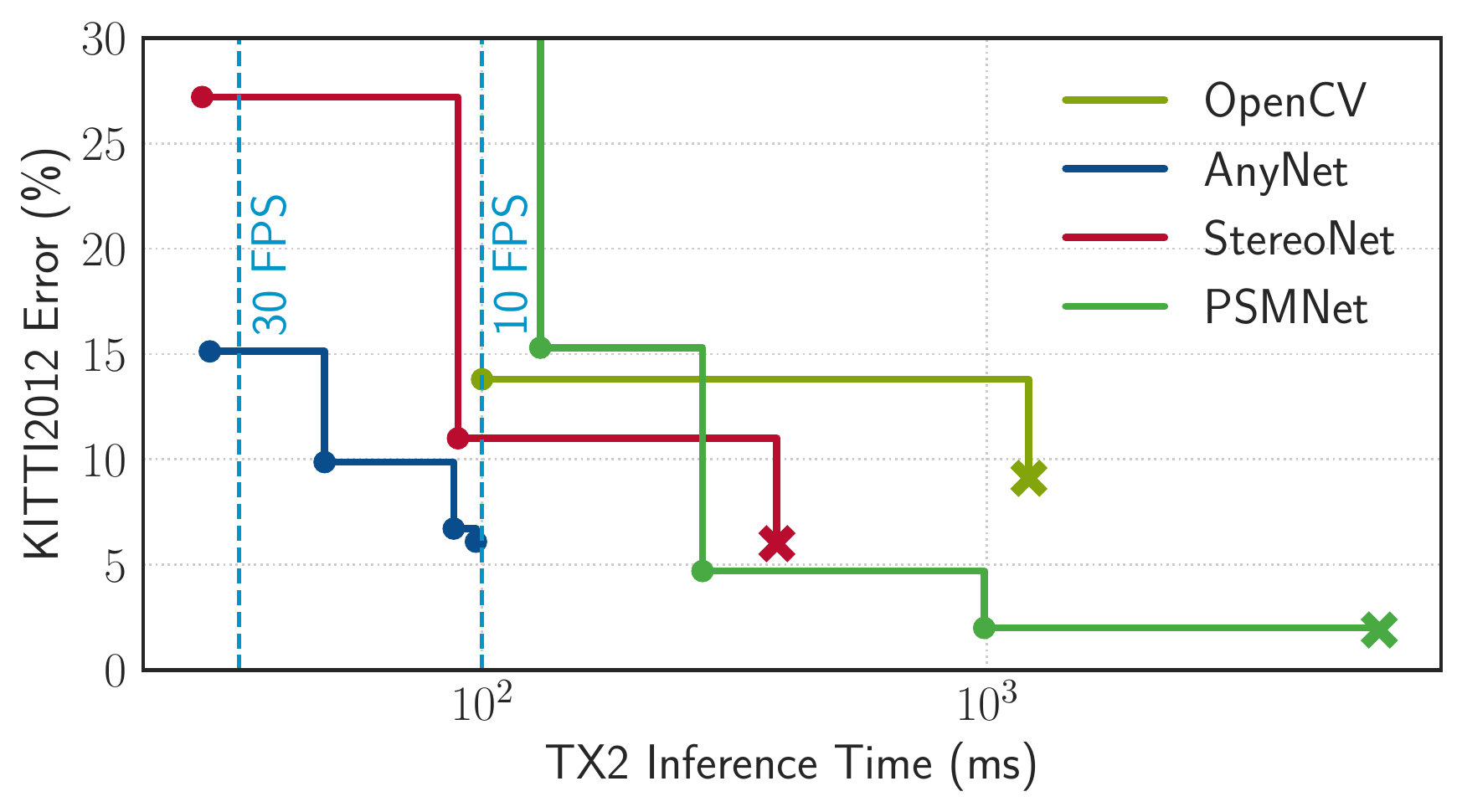}

  \caption{KITTI-2012 results using the anytime setting}\label{fig:2012_main_result_log_scale_anytime}

  \end{subfigure}\hfill
  \begin{subfigure}[t]{0.8\linewidth}

   \includegraphics[width=\linewidth ]{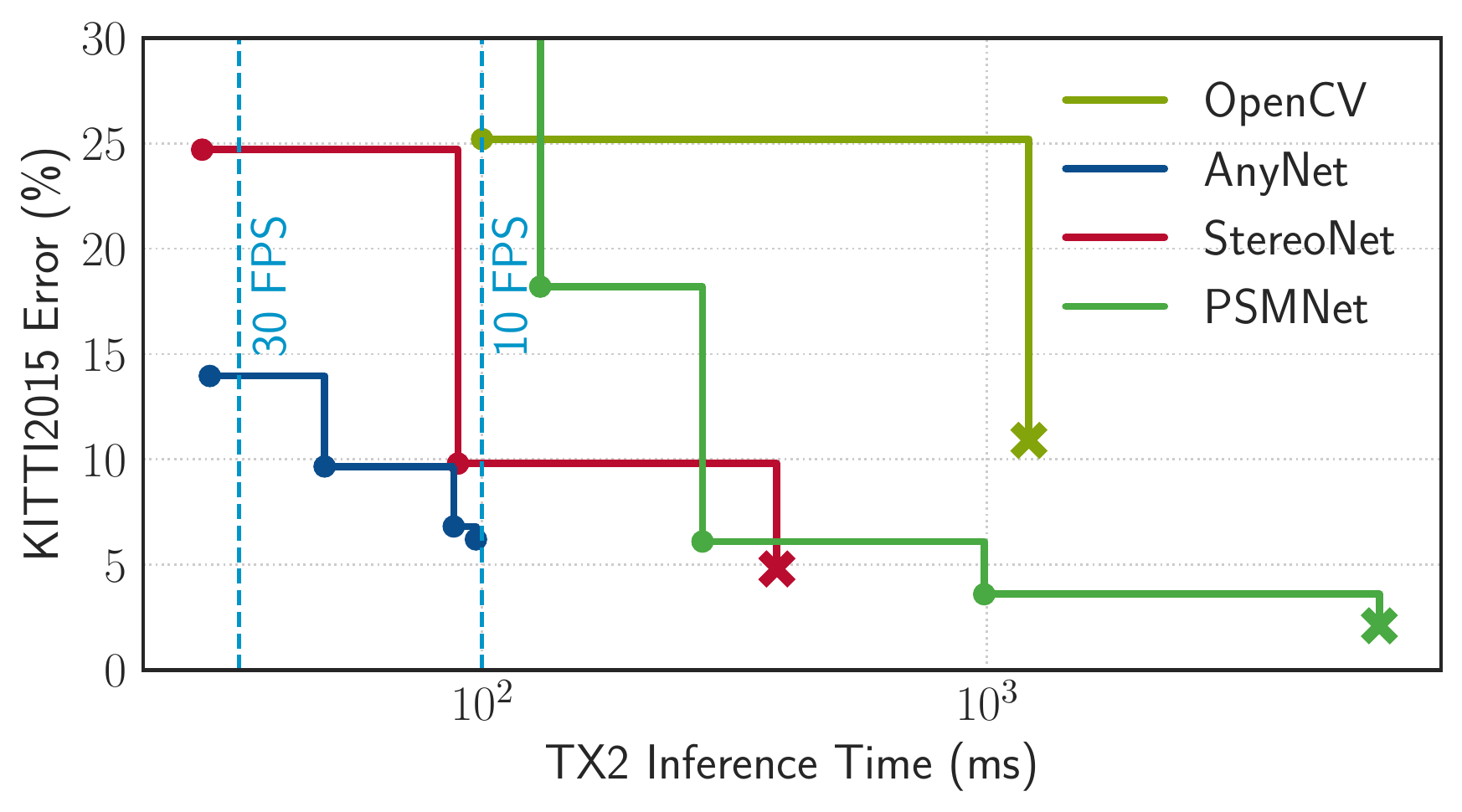}

  \caption{KITTI-2015 results using the anytime setting}\label{fig:2015_main_result_log_scale_anytime}
  \end{subfigure}
  \caption{Comparisons of the 3-pixel error rate (\%)  KITTI-2012/2015 datasets. Dots with error bars show accuracies obtained from our implementations. Crosses show values obtained from original publications.}
  \label{fig:main_results}
  \vspace{-0.5cm}
\end{figure}

Table~\ref{table:main} contains numerical results for \nameshort{} on the KITTI-2012 and KITTI-2015 datasets. Additionally, Figures \ref{fig:2012_main_result_log_scale} and \ref{fig:2012_main_result_log_scale} demonstrate the evaluation error and inference time of our model as compared to baseline methods. Baseline algorithm results originally reported in \cite{bradski2000opencv,khamis2018stereonet,zhao2017pyramid,dispnet,smolyanskiy2018importance} are shown plotted with crosses. For \nameshort{} as well as the StereoNet and PSMNet baselines, computations are performed across multiple down-sampling input resolutions. Results are generated from inputs at full resolution as well as at $1/4$, $1/8$, and $1/16$ resolution, with lower resolution corresponding to faster inference time as shown on Figs. \ref{fig:2012_main_result_log_scale} and \ref{fig:2015_main_result_log_scale}.
As seen in both plots, only \nameshort{} and StereoNet are capable of rapid real-time prediction at $\geq$30 FPS, and \nameshort{} obtains a drastically lower error rate on both data sets. \nameshort{} is additionally capable of running at over 10 FPS even with full-resolution inputs, and at each possible inference time range, \nameshort{} clearly dominates all baselines in terms of prediction error. PSMNet is capable of producing the most accurate results overall, however this is only true at computation rates of 1 FPS or slower. We also observe that the only non-CNN based approach, OpenCV, is not competitive in any inference time range.


\paragraph{\textbf{Anytime setting}}
We also evaluate \nameshort{} in the \emph{anytime} setting, in which we can poll the model prematurely at any given time $t$ in order to retrieve its most recent prediction. In order to mimic an anytime setting for the baseline OpenCV, StereoNet, and PSMNet models, we make predictions successively at increasingly higher input resolutions and execute them sequentially in ascending order of size. At time $t$ we evaluate the most recently computed disparity map. Figures \ref{fig:2012_main_result_log_scale_anytime} and \ref{fig:2015_main_result_log_scale_anytime} show the three-pixel error rates in the anytime setting. Similarly to the non-anytime results, \nameshort{} obtains significantly more accurate results in the 10-30 FPS range. Furthermore, the times between disparity map completions (visualized as horizontal lines in Figs. \ref{fig:2012_main_result_log_scale_anytime} and \ref{fig:2015_main_result_log_scale_anytime}) are much shorter than for any of the baselines, reducing the amount of wasted computation if a query is issued during a disparity map computation.

\begin{figure}
  \centering
\vspace{-0.5cm}
 \includegraphics[width=0.8\linewidth ]{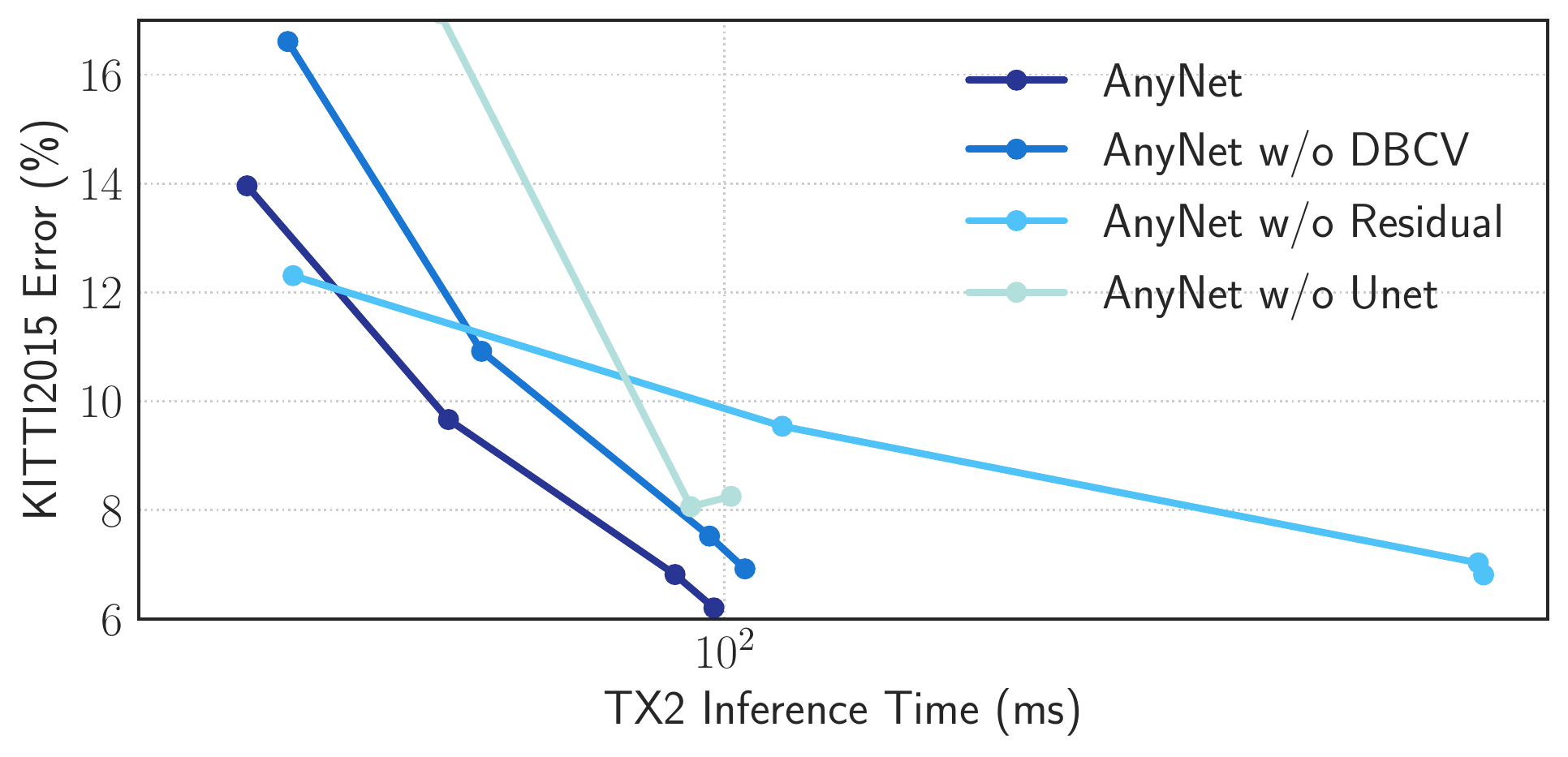}

\caption{Ablation results as three pixel error on KITTI-2015. }
  \label{fig:ablation}
\vspace{-0.5cm}
\end{figure}

\begin{table}
\begin{tabular}{M{1.1cm}M{1.35cm}M{1.35cm}M{1.35cm}M{1.55cm}}
\hline
Dataset &Stage 1 28.9ms &Stage 2  48.8ms &   Stage 3   87.9ms &  Stage 4    97.3ms\\ \hline
 KITTI2012  & $15.1 \pm 1.1$  & $9.9 \pm 0.6$ &  $ 6.7 \pm 0.4$  &  $ 6.1 \pm 0.3$      \\
 KITTI2015  & $14.0 \pm 0.7$  & $9.7 \pm 0.7$ &  $ 6.8 \pm 0.6$  &  $ 6.2 \pm 0.6$     \\ \hline

\end{tabular}

\caption{Three-Pixel error (\%) of \nameshort{} on KITTI-2012 and KITTI-2015 datasets. Lower values are better.}
\label{table:main}
  \vspace{-0.5cm}

\end{table}

\subsection{Ablation Study}
In order to examine the impact of various components of the \nameshort{} architecture, we conduct an ablation study using three variants of our model. The first replaces the U-Net feature extractor with three separated ConvNets without shared weights; the second computes a full-scale prediction at each resolution level, instead of only predicting the residual disparity; while the third replaces the distance-based cost volume construction method with the method in PSMNet~\cite{zhao2017pyramid} that produces a stack of $2 \times M$ cost volumes. All ablated variants of our method are trained from scratch, and results from evaluating them on KITTI-2015 are shown in Fig. \ref{fig:ablation}.

\paragraph{\textbf{Feature extractor}}
We modify the model's feature extractor by replacing the U-Net with three separate 2D convolutional neural networks which are similar to one another in terms of computational cost. As seen in Fig. \ref{fig:ablation} (line \nameshort{} w/o UNet), the errors increase drastically in the first two stages ($20.4\%$ and $7.3\%$). We hypothesize that by extracting contextual information from higher resolutions, the U-Net produces high-quality cost volumes even at low resolutions. This makes it a desirable choice for feature extraction.

\paragraph{\textbf{Residual Prediction}}
We compare our default network with a variant that refines the disparity estimation by directly predicting disparities, instead of residuals, in the second and third stages. Results are shown in Fig. \ref{fig:ablation} (line \nameshort{} w/o Residual). While this variant is capable of attaining similar accuracy to the original model, the evaluation time in the last two stages is increased by a factor of more than six. This increase suggests that the proposed method to predict residuals is highly efficient at refining coarse disparity maps, by avoiding the construction of large cost volumes which need to account for a large range of disparities.

\paragraph{\textbf{Distance-based Cost Volume}}
Finally, we evaluate the distance-based method for cost volume construction, by comparing it to the method used in PSMNet~\cite{zhao2017pyramid}. This method builds multiple cost volumes without explicitly calculating the distance between features from the left and right images.  The results in Fig. \ref{fig:ablation} (line \nameshort{} w/o DBCV) show that our distance-based approach is about $10\%$ faster than this choice, indicating that explicitly considering the feature distance leads to a better trade-off between accuracy and speed.

\section{Discussion and Conclusion}
  
To the best of our knowledge, \nameshort{} is the first algorithm for anytime depth estimation from stereo images. As (low-power) GPUs become more affordable and are increasingly incorporated into mobile computing devices, anytime depth estimation will enable accurate and reliable real-time depth estimation for a large variety of robotic applications. 


	
	
 \section*{ACKNOWLEDGMENT}
 This research is supported in part by grants from the National Science Foundation (III-1618134, III-1526012, IIS1149882, IIS-1724282, and TRIPODS-1740822), the Office of Naval Research DOD (N00014-17-1-2175), and the
 Bill and Melinda Gates Foundation. We are thankful for the
 generous support of SAP America Inc. We thank Camillo J. Taylor for helpful discussion.
	
	\bibliographystyle{IEEEtranS} 
	\bibliography{citations} 

\end{document}